\documentclass{article}
\usepackage[preprint]{neurips_2026}
\usepackage{iftex}
\ifPDFTeX
  \usepackage{times}
\else
  \usepackage{fontspec}
\fi
\usepackage{amsmath,amssymb}
\usepackage{booktabs}
\usepackage{graphicx}
\usepackage[hidelinks]{hyperref}
\usepackage{natbib}
\usepackage{longtable}
\usepackage{makecell}
\usepackage{listings}
\usepackage{multirow}
\usepackage{wrapfig}
\usepackage{threeparttable}
\usepackage{placeins}
\usepackage{array}
\newsavebox{\benchmarkcomparison}
\newsavebox{\benchmarkexample}
\title{InsClaimBench: Benchmarking Insurance Claim Adjudication Across the Decision Chain}

\author{%
  Linqi Zhang$^{1}$ \quad Chong Qi$^{2}$ \quad Yan Cheng$^{3}$ \quad Wanqing Cao$^{3}$ \\
  \textbf{Yu Liu$^{1}$ \quad Chenwei Lin$^{1}$\thanks{Project lead.} \quad Xian Xu$^{3}$}\thanks{Corresponding author.} \\[0.5em]
  \normalfont\small $^{1}$\,School of Computer Science and Technology, Fudan University \\
  \normalfont\small $^{2}$\,School of Integrated Circuits, Nanjing University \\
  \normalfont\small $^{3}$\,School of Economics, Fudan University
}

\date{}
\begin{document}

\maketitle

\begin{abstract}
Recent advances in reasoning-oriented large language models (LLMs) have motivated increasing evaluation of their ability to perform professional decision tasks. Insurance claim adjudication is one such task, requiring models to connect case evidence, insurance rules, intermediate judgments, and payout calculations across a structured decision process. We introduce \textbf{InsClaimBench}, an end-to-end benchmark for evaluating insurance claim adjudication across the decision chain. Grounded in real claim materials and structured insurance rules, InsClaimBench contains 3,780 cases in 375 case families across auto, property, and health insurance, comprising 86,656 atomic rule judgments. It evaluates each claim from atomic rules through adjudication modules to payout decisions and amounts, with controlled factual variants testing whether required changes are correctly propagated across levels. Evaluation of six LLMs reveals a progressive loss of reliability along the decision chain. Payout-decision accuracy ranges from 74.23--80.19\%, while joint decision--amount accuracy drops to 47.54--73.15\%. Strong local performance also fails to ensure case-level correctness: atomic-rule accuracy reaches 95.48\%, whereas rule-vector exact match peaks at only 36.90\%, and the most frequent module errors are not necessarily those most associated with final-decision failure. Under factual changes, these inconsistencies further become propagation failures: module updates are less reliable than rule updates, correct local judgments can still yield incorrect payouts, and correct payouts can conceal intermediate errors. These results show that reliable claim adjudication requires consistent composition and propagation across the decision chain.
\end{abstract}

\section{Introduction}

Recent advances in reasoning-oriented large language models (LLMs) have substantially improved their performance on complex reasoning tasks \citep{guo2025deepseek}. This progress has motivated increasing attention to LLM reasoning and decision-making in professional domains, including financial reasoning, clinical diagnosis, legal adjudication, and autonomous driving \citep{xie2026finchain,zhu2026diagnosisarena,chen2026jurisbench,ma2026drivecombo}. Such high-stakes professional decisions are rarely single predictions. They are chains of dependent judgments whose intermediate correctness and downstream consequences may diverge.

Insurance claim adjudication is one such professional decision task. It requires insurers to investigate case-specific facts, interpret applicable policy terms, determine coverage and causation, and assess the amount of loss or payment \citep{lin2024harnessing,li2025textual}. These determinations are interdependent: insurance liability depends on whether the loss falls within the insured risk and is not excluded, while the final settlement further depends on loss assessment and applicable payment conditions \citep{asmat2014does,huynh2015risk}. Whether LLMs can effectively navigate these interconnected judgments and reach appropriate claim decisions and settlement outcomes has therefore emerged as an important question for further investigation.


Existing insurance benchmarks have progressively moved beyond knowledge-oriented question answering toward reasoning tasks grounded in specific insurance scenarios and business workflows \citep{feng2015insuranceqa,chen2025inseva,lin2025insmmbench,chen2026insactbench,liu2026inslogicbench}. Despite this shift toward scenario-based decision-making, existing benchmarks typically focus on selected subtasks or partial decision processes; for example, \cite{liu2026inslogicbench} evaluate clause-grounded reasoning from case facts to claim verdicts, while \cite{lin2025insmmbench} evaluate predefined reasoning steps within specific insurance workflows. They therefore do not yet capture claim adjudication as a unified end-to-end decision process that connects underlying judgments, intermediate decisions, final claim outcomes, and their financial consequences, limiting the ability to analyze where LLMs succeed or fail along the complete decision chain.

\begin{figure}[htbp]
    \centering
    \includegraphics[width=0.85\linewidth]{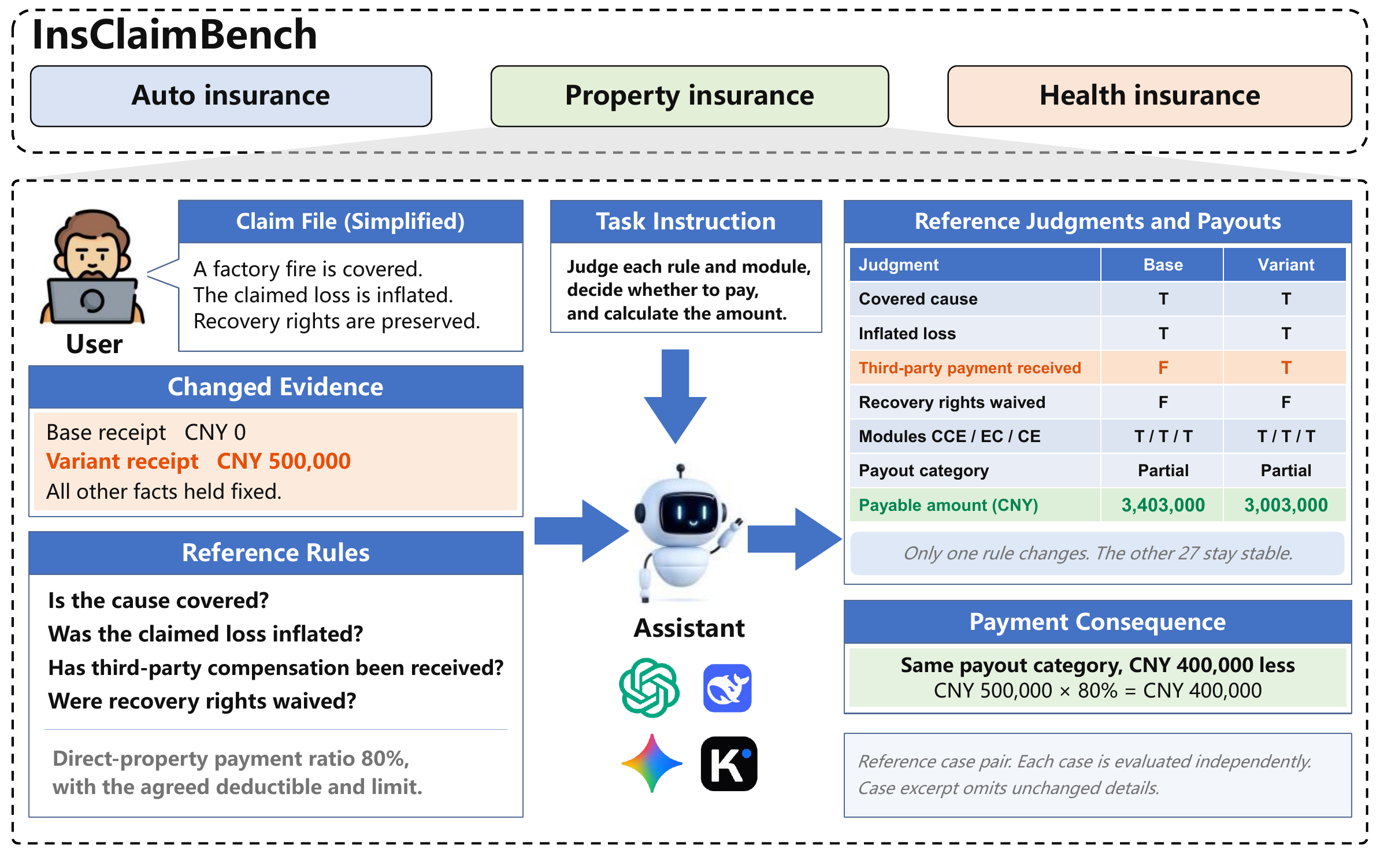}
    \caption{\textbf{Overview of InsClaimBench.} 
    An example illustrating the end-to-end claim adjudication process, including atomic rule judgments, adjudication modules, claim decisions, and payout outcomes under controlled factual changes.}
    \label{fig:benchmark-overview}
\end{figure}

To address this gap, we introduce \textbf{InsClaimBench}, an end-to-end benchmark for structured insurance claim adjudication (see Figure~\ref{fig:benchmark-overview}). InsClaimBench represents each claim as a connected decision chain spanning atomic rule judgments, adjudication modules, claim-level decisions, and payout calculation, with expert-defined logic linking the reference decisions across these levels. Grounded in real claim materials and structured insurance rules, the benchmark contains 3,780 cases organized into 375 case families across auto, property, and health insurance, comprising 86,656 atomic rule judgments. Beyond evaluating performance at individual stages, InsClaimBench constructs controlled factual variants that modify selected claim facts while holding unrelated evidence fixed, enabling systematic analysis of whether LLMs update affected judgments, preserve unaffected ones, and correctly propagate these changes to downstream claim decisions and payment outcomes.

Our evaluation of six representative LLMs reveals four main findings. First, \textbf{current LLMs do not yet achieve consistently reliable performance on insurance claim adjudication}. Claim-decision accuracy ranges from 74.23\% to 80.19\%, while joint decision-amount accuracy varies more substantially from 47.54\% to 73.15\%. Second, \textbf{strong performance on individual judgments does not reliably translate into correct end-to-end adjudication}. Atomic rule accuracy reaches as high as 95.48\%, yet the best model correctly resolves all applicable rules in only 36.90\% of cases and reaches only 73.15\% accuracy on final payout amounts. Third, \textbf{the difficulty of claim adjudication is uneven across the decision chain, with Event and Causation emerging as a consistent bottleneck}. It is the least accurate adjudication module for all six models, with accuracy ranging from 70.21\% to 75.58\%. Finally, \textbf{models exhibit substantial failures in propagating judgments through the decision chain}. Among cases requiring a module update, 28.57--39.98\% of updates fail even when the original prediction is correct; moreover, correct payment outcomes can still conceal erroneous intermediate judgments.

Our contributions are threefold: 
(1) \textbf{Decision-chain evaluation framework.} We introduce \textbf{InsClaimBench}, connecting reference rule judgments, adjudication modules, payout decisions, and amounts through expert-defined logic for end-to-end evaluation and cross-level diagnosis.
(2) \textbf{Controlled counterfactual evaluation.} We construct controlled case pairs to test required judgment updates, stability of unaffected judgments, and correctness of downstream decisions and payouts under factual changes.
(3) \textbf{Empirical diagnosis.} Across six LLMs, we show that high local accuracy does not ensure case-level correctness, the most frequent module errors need not be most associated with final-decision failure, and correct payouts can conceal intermediate errors.

\section{Related Work}

\subsection{Insurance Benchmarks}
Insurance benchmarks have gradually expanded from domain knowledge and document-based question answering to scenario-level reasoning and business-oriented tasks \citep{feng2015insuranceqa,chen2025inseva,lin2025insmmbench,liu2026inslogicbench}. This shift has introduced richer evaluation settings, including multi-step workflows, policy-grounded reasoning, and claim-level decisions. Nevertheless, most benchmarks still define their evaluation targets around individual capabilities or selected components of an insurance task, rather than the relationships among successive decisions. As a result, how performance and errors evolve from underlying judgments to final decisions and financial outcomes remains insufficiently characterized.

\subsection{Process-Level and Counterfactual Evaluation}
Beyond domain-specific benchmarks, reasoning evaluation has increasingly moved from outcome-level scoring toward examining the decision process itself. Two complementary directions have emerged: process-level evaluation localizes errors and assesses intermediate steps, rule application, or dependent task stages \citep{zheng2025processbench,song2025prmbench,xie2026finchain,zhou2025rulearena,chen2026jurisbench}, while controlled counterfactual interventions test whether model reasoning and outputs respond consistently to prescribed changes \citep{roewer2025accord,han2026rfeval}. These studies show that final-answer correctness can conceal deficiencies in intermediate reasoning and that errors may accumulate across dependent stages \citep{xie2026finchain,chen2026jurisbench,han2026rfeval}. However, the two perspectives are rarely integrated within a domain-defined professional decision process where changes in evidence have explicit implications for intermediate judgments and downstream consequences.

\section{InsClaimBench}
\label{insclaimbench}

\begin{figure}[htbp]
    \centering
    \includegraphics[width=1\linewidth]{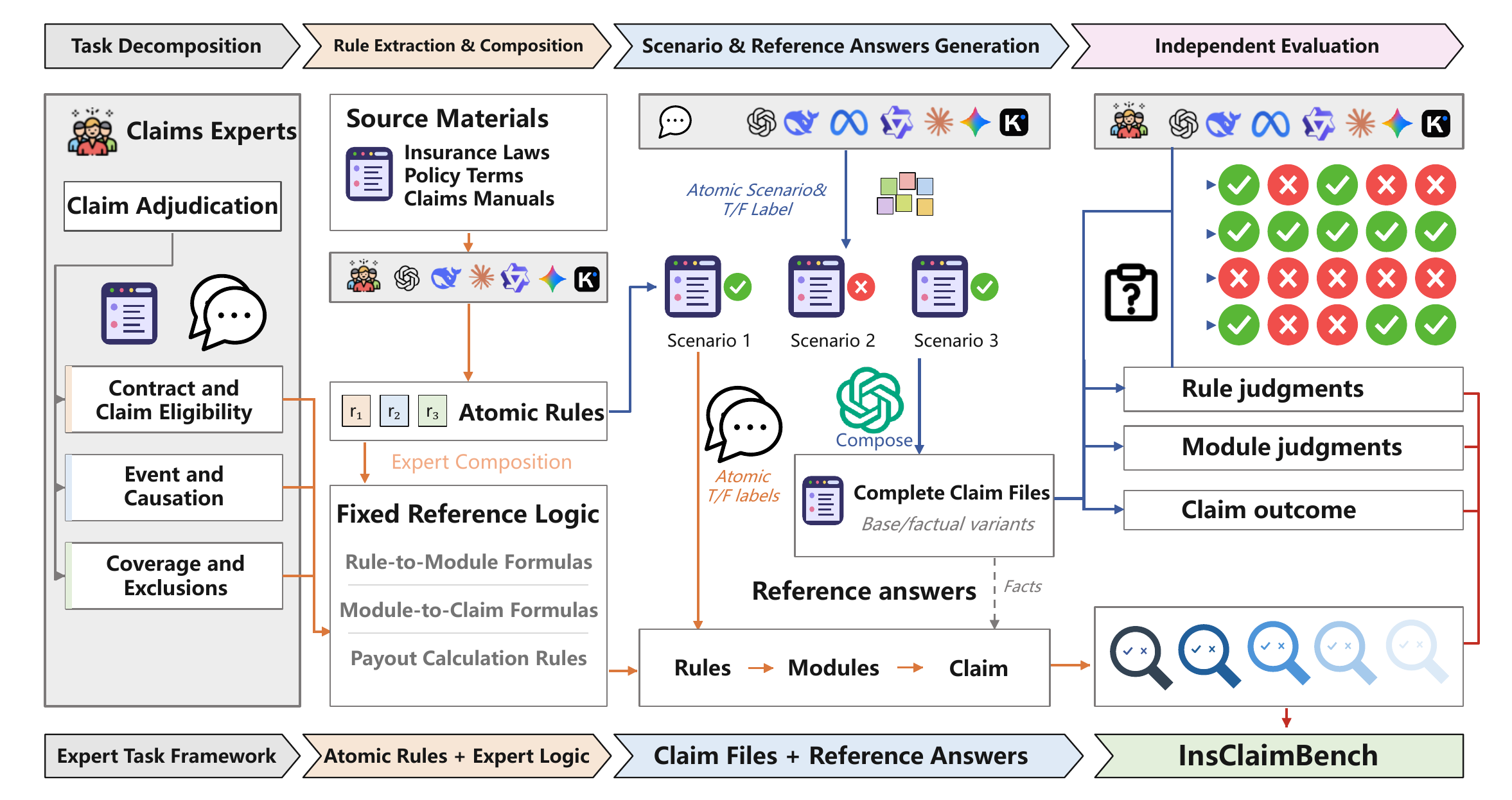}
    \caption{\textbf{Construction pipeline of InsClaimBench.} Expert-defined rules and logic guide scenario generation, reference answer construction, and independent evaluation at the rule, module, and claim levels.}
    \label{fig:task_framework}
\end{figure}

Figure~\ref{fig:task_framework} presents the overall pipeline of InsClaimBench, covering task decomposition, rule extraction and composition, case and reference-answer generation, and independent evaluation. We first define claim adjudication as a hierarchical decision task and extract atomic rules from domain materials, with expert-defined logic connecting these rules to adjudication modules, claim outcomes, and payout calculations. The resulting rules and logic guide the construction of complete claim files and controlled factual variants, after which models and human evaluators are independently assessed at different levels of the decision chain.

\subsection{Task Framework}

We define insurance claim adjudication as an end-to-end task from a complete claim file to a final payout amount. To ensure that our benchmark closely aligns with real-world insurance applications, we have developed a top-down hierarchical task definition methodology. This framework incorporates a three-layer hierarchy: 

\textbf{Claim outcome level}. Given a complete claim file, the overall task is to determine the payout decision and amount owed under the policy. The task covers auto, property, and health insurance. A predicted no-payout claim is assigned a predicted payout amount of zero in all payout-based evaluations. 

\textbf{Module level.} An insurance claim is a request for payment following a loss that may be covered under an insurance contract. Claim adjudication therefore requires determining what loss occurred and whether the policy provides coverage for that loss. These determinations can be organized into 3 legally and factually distinct modules.  \emph{Contract and Claim Eligibility (CCE)} checks whether the contract and claimant meet the requirements for a claim. \emph{Event and Causation (EC)} identifies the event, the cause of loss, and the causal chain. \emph{Coverage and Exclusions (CE)} applies coverage conditions, exclusions, and exceptions. 

\textbf{Atomic rule level.} The adjudication modules are further decomposed into atomic rules that can be judged True or False from the claim evidence. For example, a disclosure provision may require separate judgments about truthful disclosure, intentional or grossly negligent omission, its effect on underwriting, and the insurer's prior knowledge.

\begin{figure}[htbp]
    \centering
    \includegraphics[width=1\linewidth]{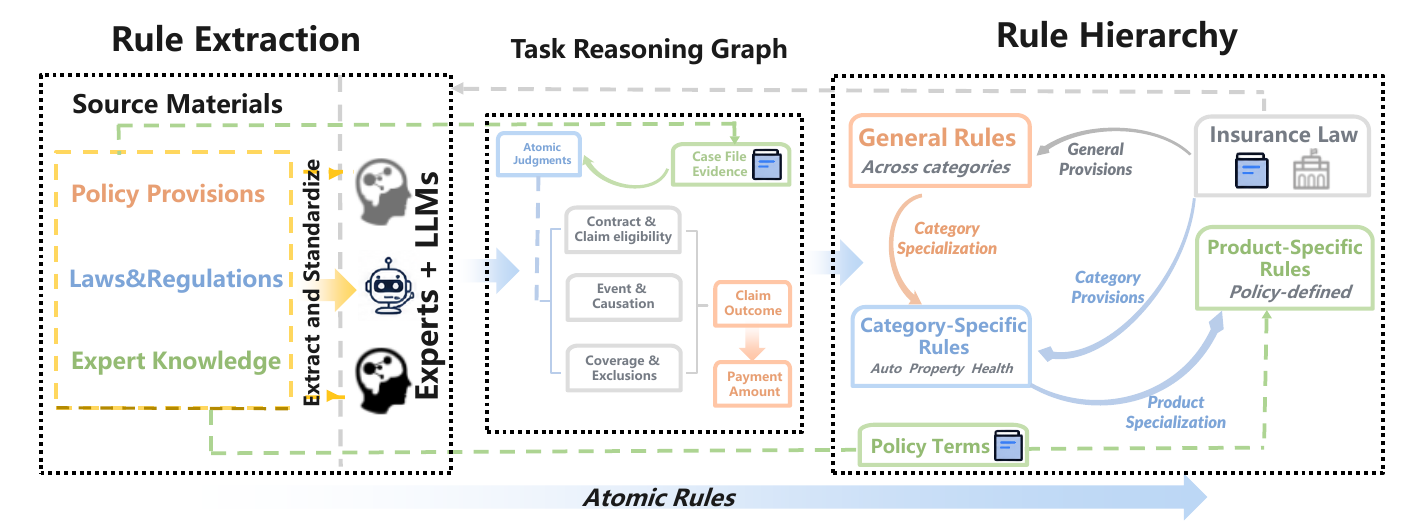}
    \caption{\textbf{Rule extraction and organization in InsClaimBench.} \textbf{Left}, claims experts and LLMs extract atomic rules from policy provisions, laws and regulations, and expert knowledge.\textbf{ Center}, the reference decision structure links case evidence and atomic judgments to adjudication modules, claim outcomes, and payout amounts. \textbf{Right}, rules are organized into general, category-specific, and product-specific levels.}
    \label{fig:rule_construction}
\end{figure}

We then define a \textbf{counterfactual pair} as a base claim file and a variant whose selected facts are modified to reverse a targeted atomic judgment, while unrelated evidence and the governing rules remain fixed. Both cases have reference answers for atomic rules, adjudication modules, payment eligibility, and final amounts. The reference logic and calculation rules determine which downstream judgments and amounts should change and which should remain unchanged.

\subsection{Dataset Construction}
\begin{table}[t]
  \centering
  \small
  \caption{Rules, rule judgments, and cases across insurance
  domains in InsClaimBench.}
  \label{tab:benchmark-scale}

  \setlength{\tabcolsep}{6pt}
  \renewcommand{\arraystretch}{1.15}

  \begin{tabular}{@{}ccccc@{}}
    \toprule
    \multicolumn{2}{c}{\makecell[c]{Insurance domain}}
      & \makecell[c]{Distinct\\rules}
      & \makecell[c]{Total rule\\judgments}
      & \makecell[c]{Cases} \\
    \midrule
    \multicolumn{2}{l}{Motor}
      & 86 & 32,054 & 1,265 \\
    \multicolumn{2}{l}{Property}
      & 28 & 35,000 & 1,250 \\
    \multirow{2}{*}{Health}
      & Critical illness & 43 & 8,701 & 627 \\
      & Medical & 48 & 10,901 & 638 \\
    \midrule
    \multicolumn{2}{l}{Total}
      & \textemdash & 86,656 & 3,780 (375) \\
    \bottomrule
  \end{tabular}
\end{table}

\noindent\textbf{Atomic rule extraction.}
Claims experts and LLMs extract atomic rules from policy terms, insurance laws and regulations, and expert knowledge (Figure~\ref{fig:rule_construction}). Each rule specifies a proposition to judge from case evidence. Rules are organized into general, category-specific, and product-specific levels according to their applicability. 

\noindent\textbf{Expert-defined formulas.}
We combine the atomic judgments into module decisions and payment eligibility. The resulting judgments and case facts determine the reference payout amount under the applicable calculation rules. To improve transparency and enable limited replication of the reference-answer construction procedure, Appendix~\ref{app:worked_example} provides a simplified and specified worked example from atomic judgments through module aggregation and payout calculation.

\noindent\textbf{Case generation.}
An LLM~(in this paper, GPT-5.6-sol) generates atomic scenarios from the extracted rules and combines them into complete claim files. Each atomic scenario supplies evidence that establishes a reference T/F label for its associated rule.

\noindent\textbf{Case family organization.}
 To construct a counterfactual variant, the generator revises the facts supporting a selected atomic judgment so that its reference label reverses, while preserving unrelated facts in the base case. For each base case and variant, we use the atomic reference labels and case facts to compute module decisions, payment eligibility, and final amounts under fixed expert logic and calculation rules. Comparing the reference answers at the two endpoints identifies the required changes and invariances at each level. We group each base case with its variants into a case family and pair every variant with its base case. The resulting dataset contains 375 families comprising 3,780 cases and 3,405 base--variant pairs. Table~\ref{tab:benchmark-scale} summarizes the distribution of rules, rule judgments, and cases across insurance domains.

\subsection{Expert Review and Human Validation}

Four claims experts, organized into two pairs, independently review a stratified sample of approximately 10\% of case families without access to the reference answers. Their atomic judgments match the reference labels with 93.99\% accuracy, with module accuracies of 93.47–97.61\%, payout decision accuracy of 96.79\%, and amount accuracy of 96.32\%~(Table~\ref{tab:human-quality}). This correspondence validates the correctness of the generated atomic scenarios and their associated answers, and the aggregation logic linking atomic judgments to downstream decisions and payouts. The two expert pairs achieve agreement rates of 91.54\% and 91.30\%, with Cohen’s $\kappa$ of 0.822 and 0.805~(Appendix~\ref{QualityVer}), supporting consistent interpretation of the generated cases and rules.

\begin{table}[htbp]
  \centering
  \small
  \caption{Human evaluation on the sampled subset (\%). Each metric is computed separately for each of the four experts on their assigned subset and then averaged across experts.}
  \label{tab:human-quality}
  \setlength{\tabcolsep}{4pt}
  \renewcommand{\arraystretch}{1.12}
  \begin{tabular}{@{}cccccccc@{}}
    \toprule
    \multicolumn{2}{c}{Rule} & \multicolumn{3}{c}{Module} & \multicolumn{3}{c}{Claim output} \\
    \cmidrule(lr){1-2}\cmidrule(lr){3-5}\cmidrule(lr){6-8}
    Acc. & Vector EM & CCE & EC & CE & T/F Acc. & Amount Acc. & MAPE \\
    \midrule
    93.99 & 92.46 & 93.47 & 97.61 & 95.13 & 96.79 & 96.32 & 2.61 \\
    \bottomrule
  \end{tabular}
\end{table}

\FloatBarrier

\section{Experiment}
\subsection{Model Selection}
We evaluate six models from different model families. These include GPT-5.6-sol from OpenAI~\citep{openai2026gpt56sol}, Gemini-3.8-Flash from Google DeepMind~\citep{google2026gemini38flash}, Qwen-3.8-Flash from Alibaba~\citep{qwen2026qwen38flash}, Kimi-K2.6 from Moonshot AI~\citep{moonshot2026kimik26}, DeepSeek-V4.1-Flash~\citep{deepseek2026v4}, and Qwen3.5-27B model. We evaluate all models on the same cases and rules. All inputs are text.

\subsection{Evaluation Protocol}
Models and human evaluators complete the adjudication in separate steps. They first judge each atomic rule, then independently assess the three modules, CCE, EC, and CE, and finally decide whether the claim is payable. For rule-level evaluation, each request contains the complete claim file and the original text of the target atomic rule. Module judgments and claim-level payout decisions are obtained through separate requests containing the complete claim file, without supplying the atomic rule texts or predictions from the rule-level evaluation. When a model judges a claim payable, a separate request to the same model uses the claim file and its directly predicted payout decision to compute the payout amount. Human evaluators likewise provide an amount when they judge a claim payable.

\subsection{Evaluation Metrics}
\label{sec:evaluation-metrics}

\noindent\textbf{Judgment scores.}
Rule accuracy pools all applicable rule judgments. Module accuracy and final T/F accuracy use all cases. \textit{Rule-vector Exact Match} (\textit{EM}) requires every rule in a case to be correct. $R^+$ and $R^-$ denote cases with all rules correct and at least one rule error. $D^+$ and $D^-$ denote correct and incorrect final decisions. The shares $R^+D^-$ and $R^-D^+$ use all cases as the denominator. \textit{Final fail.} is the percentage of wrong final decisions among cases with an error in the given module, with the error-case count in parentheses. \textit{Top rules} are ranked by error count, and their error rates use occurrences within the same subset.

\noindent\textbf{Payout amounts.}
Let $\mathcal A=\{i\mid\hat c_i=1\}$ contain cases predicted payable and $\mathcal A_+=\{i\in\mathcal A\mid q_i>0\}$. We report mean absolute percentage error (MAPE) as
\begin{equation}
\mathrm{MAPE} = \frac{100}{|\mathcal A_+|}\sum_{i\in\mathcal A_+}\frac{|\hat q_i-q_i|}{q_i}.
\end{equation}

For each base--variant pair $(x_i,x_{i'})$, we evaluate the two complete claim files independently. The reference answers define the expected changes and invariances in rule judgments, module decisions, and payout amounts.

For end-to-end evaluation, we set $\tilde q_i=\hat q_i$ for predicted payable claims and $\tilde q_i=0$ otherwise, and report
\begin{equation}
\mathrm{Acc}_{\mathrm{joint}}
=
\frac{100}{|\mathcal{J}|}
\sum_{i\in\mathcal{J}}
\mathbf{1}[\hat c_i=c_i,\,
|\tilde q_i-q_i|\le\tau_i],
\qquad
\tau_i = \max\{1,\,0.0001q_i\},
\end{equation}
where $\mathcal{J}$ contains 3,761 cases, excluding 19 cases with missing payout references.

\noindent\textbf{Local response.}
Let $z,z'$ be binary reference labels for the same rule or module, with hats denoting predictions. \emph{Sensitivity} measures whether required changes occur, and \emph{Invariance} measures whether judgments remain stable when required. Across case pairs, we define
\begin{equation}
\begin{aligned}
\mathrm{Sensitivity} &= \Pr(\hat z'\ne\hat z\mid z'\ne z),\\
\mathrm{Invariance} &= \Pr(\hat z'=\hat z\mid z'=z).
\end{aligned}
\label{eq:cf_response}
\end{equation}

\noindent\textbf{Pair correctness.}
\textit{Sensitivity} and \textit{Invariance} measure whether model predictions change when the reference label changes and remain unchanged when the reference label is unchanged. We then report pair correctness separately for changed and unchanged reference labels:
\begin{equation}
\mathrm{PC}_{\Delta}
=
\Pr(\hat z=z,\hat z'=z' \mid z'\neq z),
\qquad
\mathrm{PC}_{=}
=
\Pr(\hat z=z,\hat z'=z' \mid z'=z).
\end{equation}

Both metrics require the predictions at the two endpoints to match their corresponding references. \textit{Module Update Failure} (\textit{MUF}) counts incorrect variant judgments among pairs where the base module prediction is correct and the reference label changes. \textit{Payment Adjustment Failure} (\textit{PAF}) and \textit{Masked Module Error} (\textit{MME}) use eligible case pairs, whose counts are shown as $n$. \textit{PAF} requires at least one rule change and correct judgments for all changed rules at both endpoints. See more details in Appendix~\ref{app:metric-definitions}. All scores are percentages.

\noindent\textbf{Consequence propagation.}
We examine whether changes in rule judgments are correctly reflected in module decisions and final amounts. \emph{Payment adjustment failure} occurs when all changed rules are judged correctly and the base amount is correct, but the variant amount is wrong. \emph{Masked module error} occurs when the reference amount stays fixed and both predicted amounts are correct, yet a module whose reference label changes is judged incorrectly in either case. These checks capture errors in payment updates and intermediate judgments hidden by correct payments.

\noindent\textbf{Generator comparison.}
Differences are final T/F accuracy on DeepSeek-generated cases minus accuracy on GPT-generated cases, in percentage points. We use paired family-bootstrap 95\% confidence intervals, resampling matched case families with their base cases and variants kept together.

\section{Results}

Through experiments we find LLMs can be accurate at individual adjudication judgments, yet reliability degrades when correctness must remain coherent across the decision hierarchy and under factual changes. This degradation is heterogeneous across the adjudication chain, because frequent local errors and errors associated with downstream failures do not necessarily occur at the same stage.

\subsection{Overall Adjudication Performance}
\label{sec:overall}

\textbf{Binary payout decisions substantially overstate end-to-end adjudication reliability.} Table~\ref{tab:model-case-accuracy} shows that payout decision accuracy ranges from 74.23\% to 80.19\%, whereas joint decision--amount accuracy ranges much more widely from 47.54\% to 73.15\%. 

\textbf{Aggregate performance masks substantial variation across claim types.} As shown in Table~\ref{tab:model-case-accuracy}, payout decision accuracy on medical claims ranges from 65.05\% to 73.51\% and is lower than the corresponding accuracy on the other claim types for all six models. The variation can be substantial within the same model. For example, Kimi-K2.6 reaches 86.80\% accuracy on auto claims but 69.91\% on medical claims. Overall payout accuracy therefore conceals considerable heterogeneity across adjudication settings.

\begin{table}[htbp]
\centering
\caption{Payout decision and amount performance (\%). Decision accuracy is
shown overall and by insurance type. Amount metrics are pooled across insurance
types.}
\label{tab:model-case-accuracy}
\resizebox{\linewidth}{!}{
\begin{tabular}{lccccccc}
\toprule
& \multicolumn{5}{c}{Payout decision accuracy}
& \multicolumn{2}{c}{Payout calculation} \\
\cmidrule(lr){2-6}\cmidrule(lr){7-8}
Model & Overall & Auto & Property & Illness & Medical
& MAPE & Joint Acc. \\
\midrule
GPT-5.6-Sol         & 80.05 & 82.61 & 79.52 & 82.62 & \textbf{73.51} & \textbf{5.63} & 69.72 \\
Gemini-3.8-Flash    & 79.29 & 76.36 & \textbf{84.24} & \textbf{84.53} & 70.22 & 5.82 & \textbf{73.15} \\
DeepSeek-V4.1-Flash & 77.46 & 79.84 & 74.72 & 83.89 & 71.79 & 7.67 & 69.21 \\
Qwen3.8-Flash       & 76.64 & 80.79 & 73.52 & 79.43 & 71.79 & 7.73 & 64.88 \\
Kimi-K2.6           & \textbf{80.19} & \textbf{86.80} & 78.40 & 80.86 & 69.91 & 10.45 & 64.26 \\
Qwen3.5-27B         & 74.23 & 80.40 & 72.16 & 75.28 & 65.05 & 28.76 & 47.54 \\
\bottomrule
\end{tabular}}
\end{table}

\subsection{From Local Judgments to Case-Level Decisions}

\textbf{Strong local judgment accuracy does not translate into reliable case-level adjudication.} As shown in Table~\ref{tab:merged-joint-rule}, individual rule accuracy remains high across models, yet the probability that all applicable rules are simultaneously correct drops sharply at the case level. For example, Kimi-K2.6 reaches 89.31\% rule accuracy but only 14.68\% exact match. High local accuracy therefore still permits substantial error accumulation within complete claims.

\textbf{Local rule correctness and final payout decisions diverge in both directions.} Across models, 1.98\%--5.05\% of cases have all applicable rules judged correctly but an incorrect final payout decision, while 48.20\%--67.49\% contain at least one incorrect rule judgment despite a correct final decision (Table~\ref{tab:merged-joint-rule}). Final decision accuracy can therefore conceal substantial inconsistency in the underlying adjudication process.

\begin{table}[htbp]
\centering
\small
\caption{Rule accuracy and agreement with final T/F decisions (\%). Metrics follow Section~\ref{sec:evaluation-metrics}.}
\label{tab:merged-joint-rule}
\setlength{\tabcolsep}{4pt}
\renewcommand{\arraystretch}{1.12}
\begin{tabular}{@{}lcccc@{}}
\toprule
Model & Rule Acc. & \textbf{Vector EM} & $R^+D^-$ & $R^-D^+$ \\
\midrule
GPT-5.6-Sol & \textbf{95.48} & \textbf{36.90} & 5.05 & 48.20 \\
Gemini-3.8-Flash & 94.58 & 31.96 & 3.70 & 51.03 \\
DeepSeek-V4.1-Flash & 87.13 & 26.51 & 3.68 & 54.63 \\
Qwen3.8-Flash & 91.85 & 26.43 & 3.94 & 54.15 \\
Kimi-K2.6 & 89.31 & 14.68 & 1.98 & 67.49 \\
Qwen3.5-27B & 91.48 & 15.03 & 2.65 & 61.85 \\
\bottomrule
\end{tabular}
\end{table}

\subsection{Error Patterns Across Adjudication Modules}
\label{sec:rule-results}

\textbf{The most frequent module errors are not the errors most strongly associated with final-decision failure.} As shown in Table~\ref{tab:model-layered-metrics}, \textit{Event and Causation} is consistently the least accurate module and accounts for the largest number of module errors across all six models, while \textit{Contract and Claim Eligibility} and \textit{Coverage and Exclusions} achieve substantially higher accuracy.

\textbf{Errors in different adjudication modules have different relationships with final decision failure.} Despite being the most frequent, \textit{Event and Causation} errors are generally less often accompanied by an incorrect final decision than errors in the other two modules. For GPT-5.6-Sol, final-decision failure occurs in 42.25\% of cases with an \textit{Event and Causation} error, compared with 62.63\% for \textit{Contract and Claim Eligibility} and 59.05\% for \textit{Coverage and Exclusions}. Module accuracy and downstream failure therefore capture different aspects of adjudication reliability. These static associations do not reveal whether models update the adjudication chain consistently when the underlying evidence changes. We therefore turn to controlled factual variants.
\begin{table}[htbp]
\centering
\small
\caption{Module accuracy and final T/F failure when a module is wrong (\%). Error-case counts are in parentheses. Metrics follow Section~\ref{sec:evaluation-metrics}.}
\label{tab:model-layered-metrics}
\setlength{\tabcolsep}{4pt}
\renewcommand{\arraystretch}{1.12}
\begin{tabular}{@{}lcccccc@{}}
\toprule
\label{tab:module-error-final-failure}
\multirow{2}{*}{Model} & \multicolumn{2}{c}{CCE} & \multicolumn{2}{c}{EC} & \multicolumn{2}{c}{CE} \\
\cmidrule(lr){2-3}\cmidrule(lr){4-5}\cmidrule(lr){6-7}
 & Acc. & Final fail. ($n$) & Acc. & Final fail. ($n$) & Acc. & Final fail. ($n$) \\
\midrule
GPT-5.6-Sol & \textbf{92.57} & 62.63 (281) & 73.39 & 42.25 (1,006) & 91.67 & 59.05 (315) \\
Gemini-3.8-Flash & 86.06 & 51.42 (527) & \textbf{75.58} & 41.60 (923) & \textbf{92.22} & 48.98 (294) \\
DeepSeek-V4.1-Flash & 78.68 & 40.94 (806) & 70.21 & 29.13 (1,126) & 83.57 & 42.83 (621) \\
Qwen3.8-Flash & 83.36 & 48.97 (629) & 73.49 & 38.62 (1,002) & 87.49 & 53.70 (473) \\
Kimi-K2.6 & 80.74 & 34.07 (728) & 71.14 & 27.68 (1,091) & 88.07 & 52.55 (451) \\
Qwen3.5-27B & 89.37 & 36.07 (402) & 70.85 & 42.20 (1,102) & 84.26 & 71.26 (595) \\
\bottomrule
\end{tabular}

\end{table}

\FloatBarrier
\subsection{Decision Propagation under Factual Changes}
\label{sec:counterfactual}

Controlled factual changes expose a progressive loss of reliability as required updates propagate through the adjudication chain, from atomic rules to module decisions and ultimately to payout outcomes (Table~\ref{tab:counterfactual}).

\textbf{Adjudication reliability weakens when factual changes must be propagated from atomic rules to module-level decisions.} As shown in Table~\ref{tab:counterfactual}, module-level ${PC}_{\Delta}$ is consistently 9.92–17.43 percentage points lower than rule-level 
${PC}_{\Delta}$ across all six models. Even when the base module prediction is correct, 28.57\%–39.98\% of required module updates are incorrect. Module judgments are also less stable when no update is required, with module-level ${PC}_{=}$ consistently below rule-level ${PC}_{=}$. This gap reflects both failures to update modules when their reference decisions change and spurious changes when they should remain stable. Stronger performance on local rule updates therefore does not consistently carry over to higher-level adjudication judgments.

\textbf{Correctly handling changed atomic rules does not guarantee a correct downstream payout amount.} Even when all changed rules are judged correctly at both endpoints and the base payout amount is correct, 8.69\%--18.89\% of eligible pairs still produce an incorrect variant amount (Table~\ref{tab:counterfactual}). 

\textbf{Correct payout outcomes can likewise conceal failures in intermediate adjudication decisions.} Among pairs with correct payout amounts at both endpoints and at least one required module change, a substantial fraction still contains an incorrect judgment in a module whose reference label changes, reaching 68.03\% for some models (Table~\ref{tab:counterfactual}). 

\begin{table}[htbp]
\centering
\small
\caption{Counterfactual responses and payout outcomes on case pairs (\%). Eligible pair counts for PAF and MME are in parentheses. Metrics follow Section~\ref{sec:evaluation-metrics}.}
\label{tab:counterfactual}
\setlength{\tabcolsep}{4pt}
\renewcommand{\arraystretch}{1.12}
\resizebox{\linewidth}{!}{%
\begin{tabular}{@{}l*{9}{r}cc@{}}
\toprule
\multirow{2}{*}{Model}
& \multicolumn{4}{c}{Rule}
& \multicolumn{5}{c}{Module}
& \multicolumn{2}{c}{Payout outcomes} \\
\cmidrule(lr){2-5}\cmidrule(lr){6-10}\cmidrule(lr){11-12}
& Sens. & Inv. & $\mathrm{PC}_{\Delta}$ & $\mathrm{PC}_{=}$
& Sens. & Inv. & $\mathrm{PC}_{\Delta}$ & $\mathrm{PC}_{=}$ & MUF
& PAF ($n$) & MME ($n$) \\
\midrule
GPT-5.6-Sol
& 84.80 & 96.40 & 84.22 & 94.32
& 67.79 & 89.68 & 67.71 & 84.67 & 29.55
& 11.25 (1,813) & 34.23 (111) \\
Gemini-3.8-Flash
& 84.04 & 97.06 & 83.75 & 93.71
& 69.08 & 90.99 & 69.08 & 84.10 & 28.57
& 9.27 (2,027) & 27.41 (135) \\
DeepSeek-V4.1-Flash
& 71.48 & 92.86 & 70.68 & 84.53
& 61.76 & 83.77 & 60.76 & 73.13 & 34.05
& 8.69 (1,415) & 68.03 (147) \\
Qwen3.8-Flash
& 76.80 & 94.27 & 76.14 & 89.84
& 60.38 & 88.60 & 59.85 & 79.93 & 36.00
& 10.81 (1,453) & 66.96 (112) \\
Kimi-K2.6
& 71.93 & 94.00 & 71.38 & 87.30
& 59.69 & 88.78 & 59.54 & 78.87 & 37.00
& 18.89 (1,212) & 49.63 (135) \\
Qwen3.5-27B
& 69.66 & 97.35 & 68.88 & 91.15
& 51.83 & 93.83 & 51.45 & 82.28 & 39.98
& 14.39 (848) & 61.54 (117) \\
\bottomrule
\end{tabular}%
}
\par\smallskip
\begin{minipage}{\linewidth}
\footnotesize
$\mathrm{PC}_{\Delta}$ and $\mathrm{PC}_{=}$ denote pair correctness
for changed and unchanged reference labels, respectively:
both endpoint predictions must be correct.
Local metrics are pooled over rule or module positions
across the same 3,405 case pairs.
\end{minipage}
\end{table}

\subsection{Robustness to Case Generator}
\label{sec:case-generation-robustness}
To assess whether our case generation procedure depends on a specific generator, we conduct an ablation study by replacing GPT-5.6-Sol with DeepSeek-V4-pro. We evaluate the same six models on both sets of generated cases across four insurance types. As shown in Table~\ref{tab:scenario-reference-case-acc-property}, changing the generator leads to only small variations in payout decision accuracy. Across all 24 comparisons, the absolute differences are within 1.09 percentage points, ranging from $-0.95$ to $+1.09$ points, and all paired family-bootstrap 95\% confidence intervals include zero. These results indicate that the evaluation outcomes remain stable across the two case generators, supporting the robustness of our case generation procedure.

\section{Discussion and Conclusion}

InsClaimBench evaluates insurance claim adjudication across atomic rule judgments, adjudication modules, payout decisions, and amounts, using expert-defined reference logic and controlled factual variants. Across six LLMs, our results reveal a gap between binary decision accuracy and joint decision–amount accuracy. High local judgment accuracy does not ensure correct case-level adjudication, while module error frequency and association with final-decision failure reveal different weaknesses. Under factual changes, module judgments are less reliable than atomic rule judgments. Correct judgments on changed rules can still accompany incorrect variant payouts, while correct payouts can conceal intermediate errors. These findings demonstrate the importance of evaluating judgment correctness, responses to factual changes, and payment outcomes together across the decision chain.

However, InsClaimBench is currently based on synthetic claim files, which allows precise control over factual changes and reference answers but does not fully capture the ambiguity of real-world claim documentation. Original claim files may contain missing, inconsistent, or conflicting evidence, making adjudication substantially less controlled. Future work can therefore extend the benchmark to real claim documents and examine whether the judgment-update failures and gaps between decision correctness and payment correctness observed here persist under more realistic uncertainty.

\newpage


\newpage
\bibliographystyle{iclr2027_conference}
\bibliography{iclr2025_conference}

\newpage
\appendix
\section{Expert Quality Verification}
\label{QualityVer}

To independently verify the quality of the benchmark references, we conduct a blinded expert evaluation. We stratify case families by insurance type and randomly sample approximately 10\% of the families, while retaining the complete base--variant structure within each sampled family. The sampled families are divided into two disjoint subsets, each covering approximately 5\% of the benchmark.

Four claims experts are organized into two pairs. The two experts within each pair independently evaluate the same subset. They receive the complete claim files, original atomic rules, and applicable calculation rules, but do not have access to the benchmark reference labels, rationales, module decisions, payout decisions, or reference amounts. For each sampled case, the experts judge all applicable atomic rules, directly determine \textit{CCE},\textit{ EC}, and \textit{CE}, decide payment eligibility, and calculate the payout amount when applicable.

For each pair, we measure inter-expert agreement on binary judgments using raw agreement and Cohen's $\kappa$. As shown in Table~\ref{tab:expert-agreement}, the two expert pairs achieve agreement rates of 91.54\% and 91.30\%, with Cohen's $\kappa$ values of 0.822 and 0.805, respectively. These results indicate a high degree of independent reproducibility in the expert judgments.

Raw agreement is the proportion of paired judgments on which the two experts give the same label, while Cohen's $\kappa$ additionally adjusts for agreement expected from the marginal label distributions. Regarding compensation, the human experts were compensated at a rate equivalent to 35 days of their regular pay.

\begin{table}[htbp]
\centering
\caption{Distribution of expert judgments in human validation.}
\label{tab:expert-distribution}
\setlength{\tabcolsep}{8pt}
\renewcommand{\arraystretch}{1.15}

\begin{tabular}{ccc}
\toprule
Expert & T & F \\
\midrule
A & 3,212 & 2,204 \\
B & 3,418 & 1,998 \\
C & 3,300 & 1,791 \\
D & 3,477 & 1,614 \\
\bottomrule
\end{tabular}
\end{table}

\begin{table}[htbp]
\centering
\caption{Inter-expert agreement in blinded expert quality verification.}
\label{tab:expert-agreement}
\setlength{\tabcolsep}{6pt}
\renewcommand{\arraystretch}{1.15}

\begin{tabular}{@{}ccccc@{\hspace{0.5cm}}ccccc@{}}
\toprule
\multicolumn{5}{c}{\textbf{Group 1}}
&
\multicolumn{5}{c}{\textbf{Group 2}}
\\
\cmidrule(lr){1-5}
\cmidrule(lr){6-10}

& A:T & A:F & \shortstack{Agreement(\%)} & $\kappa$
&
& C:T & C:F & \shortstack{Agreement(\%)} & $\kappa$
\\
\midrule

B:T & 3,086 & 332
& \multirow{2}{*}{91.54}
& \multirow{2}{*}{0.822}
&
D:T & 3,167 & 310
& \multirow{2}{*}{91.30}
& \multirow{2}{*}{0.805}
\\

B:F & 126 & 1,872
& &
&
D:F & 133 & 1,481
& &
\\

\bottomrule
\end{tabular}
\end{table}
\newpage
\section{More Results}
\subsection{Top Rule Errors}

\begin{table}[htbp]
\centering
\small
\caption{Frequent rule errors (\%). Top rules are ranked by error count within each subset. All listed rules are from property insurance. Metrics follow Section~\ref{sec:evaluation-metrics}.}
\label{tab:top-rule-errors}
\setlength{\tabcolsep}{4pt}
\renewcommand{\arraystretch}{1.12}
\begin{tabular}{@{}lcc@{}}
\toprule
\multirow{2}{*}{Model} & \multicolumn{2}{c}{Top rule (error rate)} \\
\cmidrule(lr){2-3}
 & All cases & $R^-D^+$ \\
\midrule
GPT-5.6-Sol & C07 (44.00) & C07 (43.91) \\
Gemini-3.8-Flash & C07 (39.36) & P07 (44.89) \\
DeepSeek-V4.1-Flash & P01 (76.56) & P01 (77.11) \\
Qwen3.8-Flash & P01 (75.84) & P01 (75.25) \\
Kimi-K2.6 & P21 (63.52) & P21 (68.00) \\
Qwen3.5-27B & C07 (55.60) & C07 (50.35) \\
\bottomrule
\end{tabular}
\end{table}

Table~\ref{tab:top-rule-errors} further shows that C07, which evaluates whether an expense falls under a contractual exclusion, is among the most persistent errors in property claims. For GPT-5.6-Sol, C07 is incorrect in 44.00\% of all relevant property cases and remains incorrect in 43.91\% of cases where rule errors coexist with a correct final decision. Correct case-level outcomes therefore do not imply that recurring local reasoning errors have been resolved.

\subsection{Result of Ablation Study}
\begin{table}[htbp]
\centering
\small
\caption{Case-generator ablation. Entries are final T/F accuracy differences (DeepSeek minus GPT, percentage points), with paired family-bootstrap 95\% confidence intervals.}
\label{tab:scenario-reference-case-acc-property}
\setlength{\tabcolsep}{4pt}
\renewcommand{\arraystretch}{1.12}
\begin{tabular}{@{}llcccccc@{}}
\toprule
\multicolumn{2}{l}{Insurance type} & \shortstack{GPT-5.6\\Sol} & \shortstack{Gemini-3.8\\Flash} & \shortstack{DeepSeek\\Flash} & \shortstack{Qwen3.8\\Flash} & Kimi-K2.6 & \shortstack{Qwen3.5-27B} \\
\midrule
\multicolumn{2}{l}{Auto} & \shortstack{+0.00 \\ {\scriptsize [-1.11, +1.03]}} & \shortstack{-0.95 \\ {\scriptsize [-2.06, +0.16]}} & \shortstack{-0.79 \\ {\scriptsize [-1.90, +0.24]}} & \shortstack{+0.79 \\ {\scriptsize [-0.24, +1.90]}} & \shortstack{+0.16 \\ {\scriptsize [-0.95, +1.26]}} & \shortstack{-0.47 \\ {\scriptsize [-1.66, +0.71]}} \\
\multicolumn{2}{l}{Property} & \shortstack{+0.48 \\ {\scriptsize [-0.50, +0.75]}} & \shortstack{-0.24 \\ {\scriptsize [-0.78, +0.20]}} & \shortstack{-0.16 \\ {\scriptsize [-0.59, +0.09]}} & \shortstack{-0.08 \\ {\scriptsize [-0.57, +0.42]}} & \shortstack{-0.24 \\ {\scriptsize [-0.88, +0.81]}} & \shortstack{+0.24 \\ {\scriptsize [-0.33, +0.66]}} \\
\multirow{2}{*}{Health} & Illness & \shortstack{+0.77 \\ {\scriptsize [-0.67, +2.23]}} & \shortstack{+0.14 \\ {\scriptsize [-1.28, +1.44]}} & \shortstack{+1.09 \\ {\scriptsize [-0.64, +2.85]}} & \shortstack{+1.09 \\ {\scriptsize [-0.48, +2.68]}} & \shortstack{-0.51 \\ {\scriptsize [-2.07, +0.93]}} & \shortstack{-0.52 \\ {\scriptsize [-2.07, +0.92]}} \\
 & Medical & \shortstack{-0.41 \\ {\scriptsize [-1.50, +0.69]}} & \shortstack{-0.12 \\ {\scriptsize [-1.23, +0.98]}} & \shortstack{+0.21 \\ {\scriptsize [-0.88, +1.30]}} & \shortstack{-0.43 \\ {\scriptsize [-1.52, +0.67]}} & \shortstack{-0.44 \\ {\scriptsize [-1.53, +0.65]}} & \shortstack{-0.17 \\ {\scriptsize [-1.27, +0.94]}} \\
\bottomrule
\end{tabular}

\end{table}

\newpage
\section{Formal Definitions of Evaluation Metrics}
\label{app:metric-definitions}

\paragraph{Notation for counterfactual pairs.}
Let $\mathcal{P}$ denote the set of base--variant case pairs.
For each pair $p \in \mathcal{P}$, unprimed and primed quantities
refer to the base and variant cases, respectively.
Let $r_{pj}$ and $r'_{pj}$ denote the binary reference labels
for rule $j$, and let $m_{pk}$ and $m'_{pk}$ denote the binary
reference labels for module
$k \in \mathcal{K}=\{\mathrm{CCE},\mathrm{EC},\mathrm{CE}\}$.
Hats denote the corresponding predictions.
Rules are matched across the two cases by rule identity.

The sets of rules and modules whose reference labels change are
\begin{equation}
    \Delta^{R}_{p}
    =
    \{j \mid r'_{pj} \neq r_{pj}\},
    \qquad
    \Delta^{M}_{p}
    =
    \{k \in \mathcal{K} \mid m'_{pk} \neq m_{pk}\}.
    \label{eq:changed-reference-sets}
\end{equation}

Let $q_p,q'_p$ denote the reference amounts and
$\hat{q}_p,\hat{q}'_p$ the predicted amounts used in the
pairwise amount evaluation. Amount correctness at the two
endpoints is defined separately using the corresponding
reference amount:
\begin{equation}
    \begin{aligned}
        C_p
        &=
        \mathbf{1}
        \bigl\{
            |\hat{q}_p-q_p|
            \leq \max\{1,\,0.0001q_p\}
        \bigr\}, \\
        C'_p
        &=
        \mathbf{1}
        \bigl\{
            |\hat{q}'_p-q'_p|
            \leq \max\{1,\,0.0001q'_p\}
        \bigr\}.
    \end{aligned}
    \label{eq:pair-amount-correctness}
\end{equation}

All probabilities below are empirical proportions over the
specified evaluation units and are multiplied by $100$ to
obtain percentages.

\paragraph{Module Update Failure (MUF).}
MUF measures how often a variant module prediction is incorrect
when the base prediction is correct and the reference module
label changes:
\begin{equation}
    \mathrm{MUF}
    =
    100\,
    \widehat{\Pr}_{(p,k)}
    \left(
        \hat{m}'_{pk} \neq m'_{pk}
        \,\middle|\,
        \hat{m}_{pk}=m_{pk},
        \ m'_{pk}\neq m_{pk}
    \right).
    \label{eq:muf}
\end{equation}
The evaluation unit is a pair--module position $(p,k)$.
The denominator pools all positions with a correct base
prediction and a changed reference label across the three
modules and all case pairs.

\paragraph{Payment Adjustment Failure (PAF).}
Let $B_p$ denote the event that at least one reference rule
label changes and every changed rule is predicted correctly
at both endpoints:
\begin{equation}
    B_p
    =
    \left\{
        \Delta^{R}_{p}\neq\varnothing
    \right\}
    \cap
    \bigcap_{j \in \Delta^{R}_{p}}
    \left\{
        \hat{r}_{pj}=r_{pj},
        \ \hat{r}'_{pj}=r'_{pj}
    \right\}.
    \label{eq:paf-rule-condition}
\end{equation}
PAF measures the frequency of an incorrect variant amount
conditional on $B_p$ and a correct base amount:
\begin{equation}
    \mathrm{PAF}
    =
    100\,
    \widehat{\Pr}_{p}
    \left(
        C'_p=0
        \,\middle|\,
        B_p,\ C_p=1
    \right).
    \label{eq:paf}
\end{equation}
The evaluation unit is a case pair. The eligible-pair count
reported alongside PAF is
\begin{equation}
    n_{\mathrm{PAF}}
    =
    \sum_{p \in \mathcal{P}}
    \mathbf{1}\{B_p \text{ holds},\ C_p=1\}.
    \label{eq:paf-count}
\end{equation}
This definition does not require $q'_p \neq q_p$.
It therefore includes failures to preserve a reference amount
that should remain unchanged, as well as failures to produce
a required amount change.

\paragraph{Masked Module Error (MME).}
MME measures how often correct amounts at both endpoints
coexist with an incorrect prediction for a module whose
reference label changes. Eligible pairs must have identical
reference amounts and at least one changed module:
\begin{equation}
    \begin{aligned}
        \mathrm{MME}
        =
        100\,
        \widehat{\Pr}_{p}
        \Bigl(
            &\exists k \in \Delta^{M}_{p},
            \ (\hat{m}_{pk}\neq m_{pk}
            \lor \hat{m}'_{pk}\neq m'_{pk}) \\
            &\Bigm|
            C_p=C'_p=1,\ q_p=q'_p,\
            \Delta^{M}_{p}\neq\varnothing
        \Bigr).
    \end{aligned}
    \label{eq:mme}
\end{equation}
The evaluation unit is a case pair, and each eligible pair
is counted once even if multiple changed modules contain errors.
The eligible-pair count reported alongside MME is
\begin{equation}
    n_{\mathrm{MME}}
    =
    \sum_{p \in \mathcal{P}}
    \mathbf{1}
    \bigl\{
        C_p=C'_p=1,\ q_p=q'_p,\
        \Delta^{M}_{p}\neq\varnothing
    \bigr\}.
    \label{eq:mme-count}
\end{equation}

\newpage
\section{Worked Example of Reference-Answer Construction}
\label{app:worked_example}
This section provides a fully specified illustrative example of deterministic reference-answer aggregation and payout calculation. 

\subsection{Illustrative Scope}

To make the construction of reference answers inspectable without releasing the complete proprietary rule inventory, we provide a fully specified worked example from a frozen illustrative medical-insurance product. The example exposes the complete computation from atomic rule labels to module judgments, claim-level payment eligibility, and the reference payout amount. All quantities and logical operations required to reproduce the example are given below.

This example is intended to demonstrate the deterministic aggregation procedure used in InsClaimBench. Product-specific rule inventories, source mappings, and the full set of expert composition formulas are not disclosed.

For the illustrative medical claim, let \(r_1,\ldots,r_{10}\) denote binary atomic judgments. A value of \(T\) means that the proposition represented by the rule is satisfied by the case evidence.

\begin{table}[htbp]
\centering
\small
\begin{tabular}{p{0.10\linewidth} p{0.82\linewidth}}
\toprule
Rule & Proposition represented by \(T\) \\
\midrule
\(r_1\) &
The claimant and policy satisfy the applicable eligibility requirements, and the relevant coverage is in force. \\
\(r_2\) &
The event falls within the insurance period or an applicable contractual extension. \\
\(r_3\) &
A valid disclosure-related bar to coverage is established. \\
\(r_4\) &
A medical event covered by the relevant benefit has occurred. \\
\(r_5\) &
The claimed medical expenses are related to the event, reasonable, and necessary. \\
\(r_6\) &
The waiting-period provision applies and the event falls within the waiting period. \\
\(r_7\) &
An accident-related exception to the waiting period applies. \\
\(r_8\) &
The policy otherwise waives the applicable waiting period. \\
\(r_9\) &
The hospital, geographic scope, and expense categories satisfy the applicable coverage requirements. \\
\(r_{10}\) &
A statutory bar that independently precludes payment is established. \\
\bottomrule
\end{tabular}
\caption{Atomic judgments used in the illustrative medical-insurance example.}
\label{tab:appendix_demo_rules}
\end{table}

For each contractual exclusion \(j\), we additionally define four binary judgments. Here, \(e_j\) indicates that the factual condition described by the exclusion is present, \(k_j\) indicates that the required relation between that condition and the claimed loss is established, \(v_j\) indicates that the exclusion is legally and contractually effective, and \(z_j\) indicates that an exception restoring coverage under that exclusion applies.

An exclusion blocks coverage only when
\begin{equation}
X_j
=
e_j \land k_j \land v_j \land \neg z_j.
\label{eq:appendix_exclusion}
\end{equation}

The three adjudication modules are then constructed as
\begin{align}
M_{\mathrm{CCE}}
&=
r_1 \land r_2 \land \neg r_3,
\\
M_{\mathrm{EC}}
&=
r_4 \land r_5
\land
(\neg r_6 \lor r_7 \lor r_8),
\\
M_{\mathrm{CE}}
&=
r_9
\land
\neg r_{10}
\land
\neg \bigvee_j X_j.
\end{align}

The reference claim-level payment decision is
\begin{equation}
c
=
M_{\mathrm{CCE}}
\land
M_{\mathrm{EC}}
\land
M_{\mathrm{CE}}.
\label{eq:appendix_claim_decision}
\end{equation}

This construction also makes explicit how exceptions operate. For example, \(r_6=T\) indicates that the event falls within an applicable waiting period. The waiting-period condition nevertheless passes when either \(r_7=T\) or \(r_8=T\). Likewise, an exclusion whose factual, relational, and validity requirements are all satisfied does not block coverage when its corresponding exception \(z_j=T\).

\subsection{Worked Logical Example}

Consider an illustrative claim with the following reference atomic judgments:
\begin{align}
&r_1=T,\quad r_2=T,\quad r_3=F,
\\
&r_4=T,\quad r_5=T,\quad r_6=T,
\\
&r_7=T,\quad r_8=F,\quad r_9=T,\quad r_{10}=F.
\end{align}

Suppose one potentially relevant exclusion is present and its atomic judgments are
\begin{equation}
e_1=T,\qquad
k_1=T,\qquad
v_1=T,\qquad
z_1=T.
\end{equation}

The corresponding exclusion indicator is therefore
\begin{equation}
X_1
=
T \land T \land T \land \neg T
=
F.
\end{equation}

The module references are obtained deterministically as
\begin{align}
M_{\mathrm{CCE}}
&=
T \land T \land \neg F
=
T,
\\
M_{\mathrm{EC}}
&=
T \land T
\land
(\neg T \lor T \lor F)
=
T,
\\
M_{\mathrm{CE}}
&=
T \land \neg F \land \neg F
=
T.
\end{align}

Hence,
\begin{equation}
c
=
T \land T \land T
=
T,
\end{equation}
and the claim proceeds to payout calculation.

This example illustrates two exception mechanisms simultaneously. The event occurs during the waiting period, but the accident-related exception preserves eligibility. A potentially applicable exclusion is also triggered at the factual level, but the exclusion-specific exception prevents that exclusion from blocking coverage.

\subsection{Worked Payout Calculation}

Suppose the submitted medical bill is \(24{,}000\). Of this amount, \(2{,}000\) falls outside the covered expense scope and \(10{,}000\) has already been matched to other compensation. The expense entering the benefit formula is therefore
\begin{equation}
x
=
24{,}000
-
2{,}000
-
10{,}000
=
12{,}000.
\end{equation}

For this frozen illustrative product, the applicable medical benefit schedule reimburses the first CNY \(10{,}000\) of \(x\) at \(50\%\) and the remaining eligible amount at \(100\%\). Thus,
\begin{equation}
G(x)
=
0.5\min(x,10{,}000)
+
\max(x-10{,}000,0).
\label{eq:appendix_payout_schedule}
\end{equation}

Substituting \(x=12{,}000\) gives
\begin{equation}
G(12{,}000)
=
0.5\times 10{,}000
+
2{,}000
=
7{,}000.
\end{equation}

Let \(H\) denote the remaining applicable policy limit. The adjudicated benefit is
\begin{equation}
P
=
\min\{G(x),H\}.
\label{eq:appendix_policy_cap}
\end{equation}

When \(H\ge 7{,}000\), the reference adjudicated amount is therefore
\begin{equation}
P=7{,}000.
\end{equation}

If only CNY \(6{,}500\) remains under the applicable limit, the same calculation instead yields
\begin{equation}
P=6{,}500.
\end{equation}

Where an advance payment \(V\) has already been made against the same adjudicated claim, the adjudicated benefit \(P\) remains unchanged and the amount transferred in the current settlement is
\begin{equation}
N
=
\max(P-V,0).
\label{eq:appendix_net_transfer}
\end{equation}

For example, if \(P=\mathrm{CNY}\ 7{,}000\) and \(V=\mathrm{CNY}\ 1{,}000\), then the current transfer is
\begin{equation}
N=6{,}000.
\end{equation}
Here $N$ denotes the reference amount $q$ used in the evaluation metrics.
\end{document}